%% file: icml2019.tex
\documentclass{article}

\usepackage{microtype}
\usepackage{graphicx}
\usepackage{booktabs} % for professional tables

\usepackage{hyperref}

\usepackage[accepted]{icml2019}

\icmltitlerunning{Projected Bayesian Neural Networks}

\usepackage{algorithm}
\usepackage{algorithmic}

\usepackage[utf8]{inputenc} % allow utf-8 input
\usepackage[T1]{fontenc}    % use 8-bit T1 fonts
\usepackage{lmodern}

\usepackage{hyperref}       % hyperlinks
\usepackage{url}            % simple URL typesetting
\usepackage{booktabs}       % professional-quality tables
\usepackage{amsfonts, bm}       % blackboard math symbols
\usepackage{nicefrac}       % compact symbols for 1/2, etc.
\usepackage{amsmath}
\usepackage{wrapfig}
\usepackage{natbib}
\graphicspath{{./figs/}}
\usepackage{color}
\usepackage{adjustbox}
\usepackage{subfig}

\input{00_prelude.tex}

\begin{document}

\twocolumn[
\icmltitle{Projected BNNs: Avoiding weight-space pathologies by learning latent representations of neural network weights}




\begin{icmlauthorlist}
\icmlauthor{Melanie F. Pradier}{to}
\icmlauthor{Weiwei Pan}{to}
\icmlauthor{Jiayu Yao}{to}
\icmlauthor{Soumya Ghosh}{goo}
\icmlauthor{Finale Doshi-Velez}{to}
\end{icmlauthorlist}

\icmlaffiliation{to}{School of Engineering and Applied Sciences, Harvard University, Cambridge, MA, USA}
\icmlaffiliation{goo}{IBM Research, Cambridge, MA, USA}

\icmlcorrespondingauthor{Melanie F. Pradier}{melanie.fpradier@gmail.com}
\icmlcorrespondingauthor{Weiwei Pan}{weiweipan@g.harvard.edu}

\icmlkeywords{Machine Learning, ICML}

\vskip 0.3in
]



\printAffiliationsAndNotice{}  % leave blank if no need to mention equal contribution

\vspace{-0.3cm}
\begin{abstract}
	As machine learning systems get widely adopted for high-stake decisions, quantifying uncertainty over predictions becomes crucial.  While modern neural networks are making remarkable gains in terms of predictive accuracy, characterizing uncertainty over the parameters of these models is challenging because of the high dimensionality and complex correlations of the network parameter space. This paper introduces a novel variational inference framework for Bayesian neural networks that (1) encodes complex distributions in high-dimensional parameter space with representations in a low-dimensional latent space, and (2) performs inference efficiently on the low-dimensional representations. Across a large array of synthetic and real-world datasets, we show that our method improves uncertainty characterization and model generalization when compared with methods that work directly in the parameter space.
\end{abstract}

\input{01_intro.tex}

\input{02_literature_review.tex}

\input{03_methods.tex}

\input{04_results.tex}

\input{05_conclusion.tex}

%\bibliographystyle{icml2019}
%\bibliography{arxiv}
\bibliographystyle{plainnat}
\bibliography{arxiv}

%%%%%%%%%%%%%%%%%%%%%%%%%%%%%%%%%%%%%%%%%%%%%%%%%%%%%%%%%%%%%%%%%%%%%%%%%%%%%%%
%%%%%%%%%%%%%%%%%%%%%%%%%%%%%%%%%%%%%%%%%%%%%%%%%%%%%%%%%%%%%%%%%%%%%%%%%%%%%%%
\newpage
\onecolumn

\section*{Appendix A: Details on Experimental Setup}

\subsection*{Generation of Toy Datasets}

The data in Figure 1 was generated by sampling points non-uniformly from a function represented by a feedforward network (with 1 hidden layer, 20 hidden nodes and RBF activation centered at 0 with length scale 1) whose weights are obtained by applying a fixed linear transform to a fixed latent vector $\z$.

The toy data in Figure 2 was generated by first sampling data points from the 4 modes shown in Figure 2d, and then fitting a feedforward network multiple times with a single hidden layer, three nodes and RBF activation function). Each cluster of weights corresponds to fitting a different set of three of the four modes in the data.

\subsection*{Baselines}

\begin{itemize}
	\item \textbf{Bayes by Backprop (BbB)}: We implemented this algorithm ourselved. Note that proj-BNN can be reduced to BbB when the projection function is the identity. We perform cross-validation of the learning rate in the range: $[0.0001,0.005,0.001,0.01,0.1]$.
	\item \textbf{Fast Geometric Ensembling (FGE)}: this method is used both, as a baseline, and in the first stage of our inference framework to find a smart initialization for Proj-BNN. In both cases, we use the following hyperparameters: a step-size of 0.001 to first reach the MAP solution, and then a cyclic learning rate between 0.01 and 0.0001. Each cycle lasts for 10 epochs.
	\item \textbf{Matrix Variate Gaussian Posteriors (MVG)}: we use the code in Tensorflow that is publicly available from the authors at: \url{https://github.com/AMLab-Amsterdam/SEVDL_MGP}. In their original formulation, the authors learn the observational noise: in order to allow for comparison across different baselines, for simplicity, we adapt their code to keep the observational noise fixed to $\sigma_y = 0.1$.
	\item \textbf{Bayesian Hypernetworks (BbH)}: we use the code in Tensorflow from the authors that is publicly available at \url{https://github.com/pawni/BayesByHypernet}. For the architecture of the hypernetwork, we cross-validate across the architectures used in the authors'paper, namely: $[[50],[64,64],[128,128]]$.
	\item \textbf{Multiplicative Normalizing Flows (MNF)}: we use the code that is publicly available at \url{https://github.com/AMLab-Amsterdam/MNF_VBNN}. We keep all default hyperparameters, except the learning rate, which we cross-validate using a grid search $[0.0005,0.001,0.005,0.01]$.
	% prior N(0.1)
\end{itemize}

\subsection*{Simulation Setup}

For each data set, we gather $R=500$ candidate weight solutions $\W_{\c}$ using Fast Geometric Ensembling on the training set. We then keep the top best 150 sets of weights in terms of RMSE in validation set.

In all the experiments, we use a random train-test-validation split of 80-10-10. All datasets are normalized in a preprocessing step to have zero mean and unit standard deviation.
For each optimization subtask in our inference framework, we perform cross-validation of the step size $\lambda_1 \in \{0.1, 0.01, 0.001, 0.0005, 0.0001 \}$ and fix the batch-size to 128.% \in \{16, 128, 512 \}$.
We use Adam as the optimizer and the joint unnormalized posterior distribution $p( \W | \mathcal{D}_{train})$ as the objective function. Optimization is performed with $L=50,000$ iterations and early stopping once the marginal log likelihood in validation set stop increasing after 30 iterations.

The proposed approach proj-BNN is implemented in Autograd, source code is publicly available at: \emph{Anonymized}. For the prediction-constrained autoencoder (the projection function), we fix the architecture to either a single layer of 20 nodes or 10,10 with two hidden layers. We use RBF activation functions, and additional input noise with variance 1 for robustness. We cross-validate across the dimensionality of the latent space $\z$ in the range $2,10,50,100$. We use 20 samples for the reparametrization trick.

We evaluate by computing the marginal test log likelihood as:
\begin{align}
\mathbb{E}_{p(\x^\star,\y^\star)}  \left[\log p(\y^\star|\x^\star,\mathcal{D})\right] =  \mathbb{E}_{p(\x^\star,\y^\star)} \left[\log \int p(\y^\star|\x^\star,\W) p(\W|\mathcal{D}) d\W \right]
\end{align}
%	In practice, initializing $\vpphisig$ to small values gave better performance, i.e., $\log \vpphisig \sim \mathcal{N}(-9,0.1)$.

%\subsection{Evaluation Metrics}
%
%We compute the marginal test log likelihood as:
%\begin{align}
% \mathbb{E}_{p(\x^\star,\y^\star)} & \left[\log p(\y^\star|\x^\star,\mathcal{D})\right] = \nonumber \\ & \mathbb{E}_{p(\x^\star,\y^\star)} \left[\log \int p(\y^\star|\x^\star,\W) p(\W|\mathcal{D}) d\W \right]
%\end{align}
%
%The following additional metrics can be found in the additional results of this Appendix:
%\begin{itemize}
%	\item Root Mean Square Error: $\mathrm{RMSE}=\sqrt{|\y-\hat{\y}|_2}$. The lower, the better.
%	\item Coverage of Test Points: this refers to the fraction of points in the test set covered by the posterior predictive between percentiles 5\% and 95\%. The closer to 0.90, the better.
%	\item Continuous Ranked Probability Score (CRPS):  this metric is used in the statistics community to assess the accuracy of probabilistic forecasts. The lower, the better.
%\end{itemize}

\section*{Appendix B: Further Results}

\subsection*{Impact of dimensionality of latent space}

\begin{figure*}[h!]
	\centering
	\includegraphics[width=0.23\textwidth]{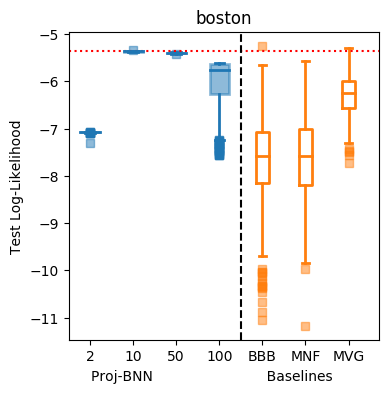}\includegraphics[width=0.23\textwidth]{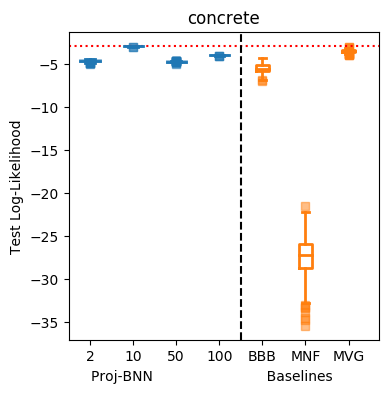}	\includegraphics[width=0.23\textwidth]{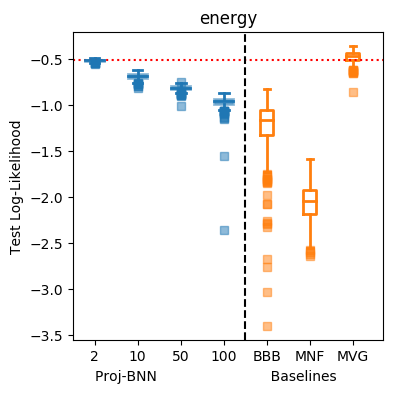}\includegraphics[width=0.23\textwidth]{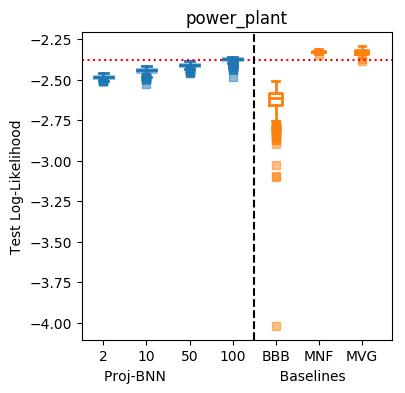}
	
	\includegraphics[width=0.23\textwidth]{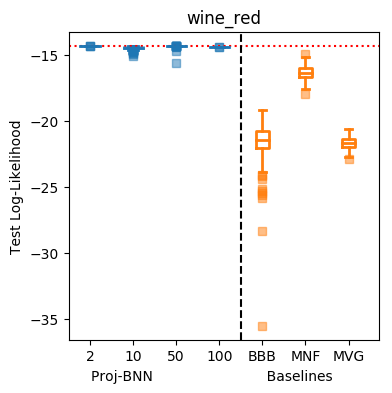}\includegraphics[width=0.23\textwidth]{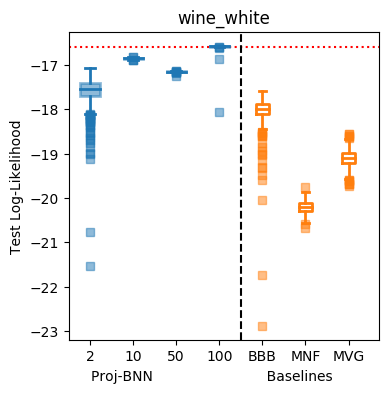}	\includegraphics[width=0.23\textwidth]{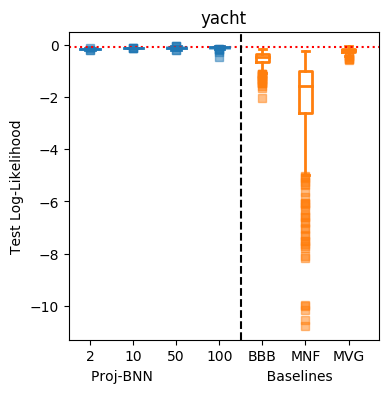}\includegraphics[width=0.23\textwidth]{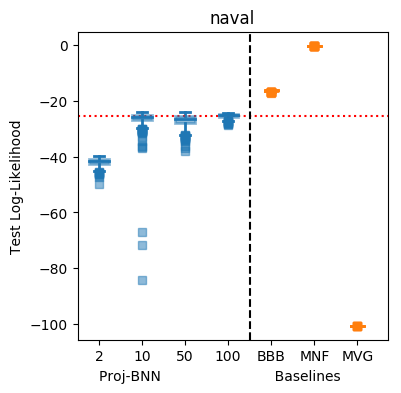}
	
	\caption{\textbf{Test log-likelihood for UCI benchmark datasets for varying dimensionality of $\z$-space.} Each dot corresponds to $p(\Y|\X,\W^{(s)})$, where $\W^{(s)}$ is one sample from the optimized variational posterior $\W^{(s)} \sim q(\W)$. Red dotted horizontal line corresponds to the best performance of our approach Proj-BNN. Baselines: a) BBB: Bayes By Backprop (Blundell, et.al 2015); b) MNF: multiplicative normalizing flow (Louizos et.al, 2017); c) MVG: multivariate Gaussian prior BNN (Louizos et.al, 2016).}
	\label{fig:var_dim_u}
\end{figure*}

Figure~\ref{fig:var_dim_u} shows the \emph{point-wise} marginal test log-likelihood $p(\Y|\X,\W^{(s)})$ for 500 different samples $\W^{(s)} \sim q(\W)$ of Proj-BNN for varying dimensionality of the latent space $\z$ across the UCI benchmark datasets considered in the paper.

\subsection*{Extrapolation and Interpolation Splits}

In this Section, we repeat the simulations on the UCI benchmark datasets but perform a splitting of the data that tests the extrapolation and interpolation capacities of the models more precisely. In particular, we split the data according to the L2 norm of the observations. We create two datasets with 80\% training, 10\% validation, and 10\% testing as follows: the extrapolation split is such that 5\% of the observations with highest norm and 5\% of the observations with lowest norm are kept in the test set; the interpolation split is the complementary to that one (that is, an inverted doughnut). Interestingly, BbB proves to be quite robust across all datasets. Overall, our approach is more robust to other structured variational inference techniques.

%%%   FIGURE
\begin{figure*}[h!]
	\includegraphics[width=0.5\textwidth]{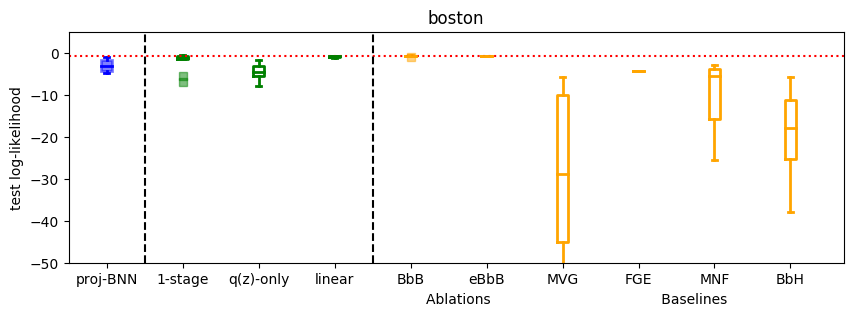}\includegraphics[width=0.5\textwidth]{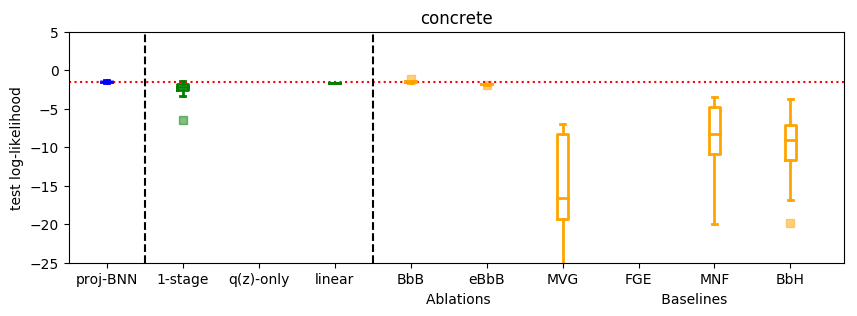}
	\includegraphics[width=0.5\textwidth]{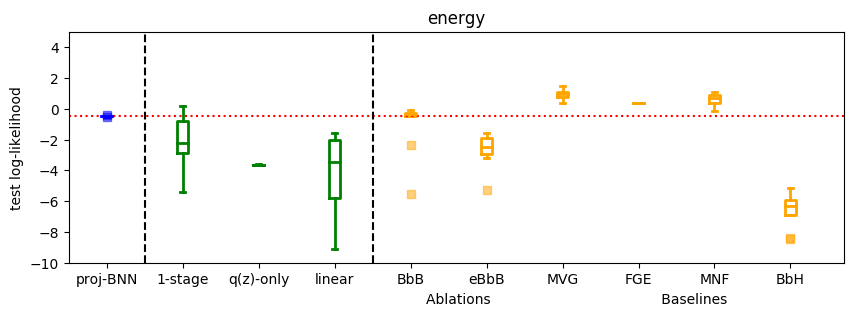}\includegraphics[width=0.5\textwidth]{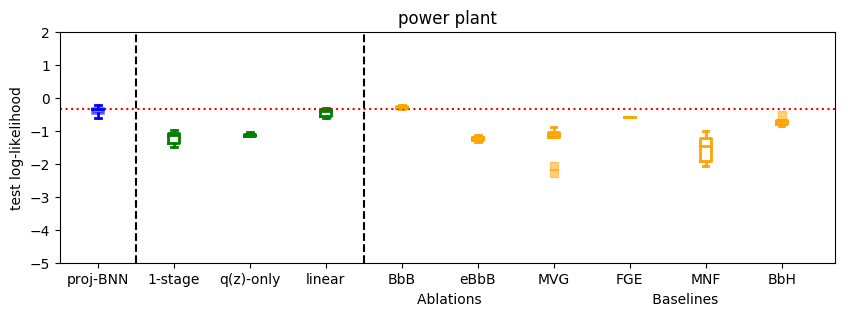}
	\includegraphics[width=0.5\textwidth]{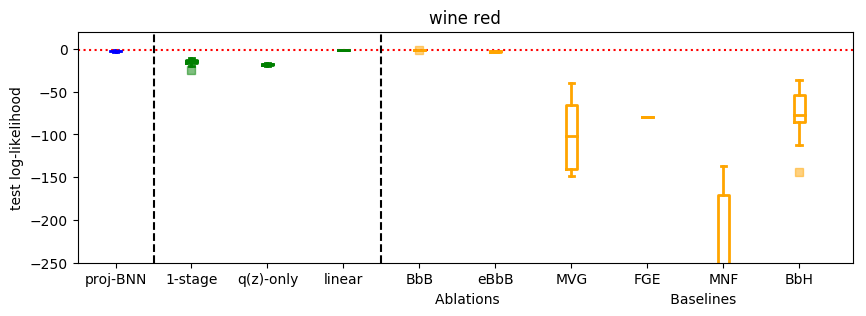}\includegraphics[width=0.5\textwidth]{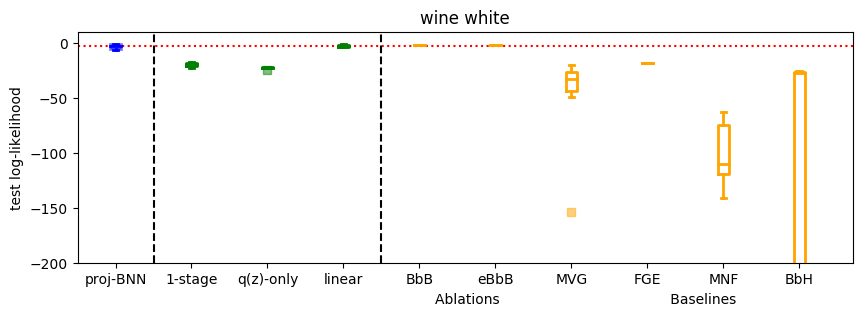}
	\includegraphics[width=0.5\textwidth]{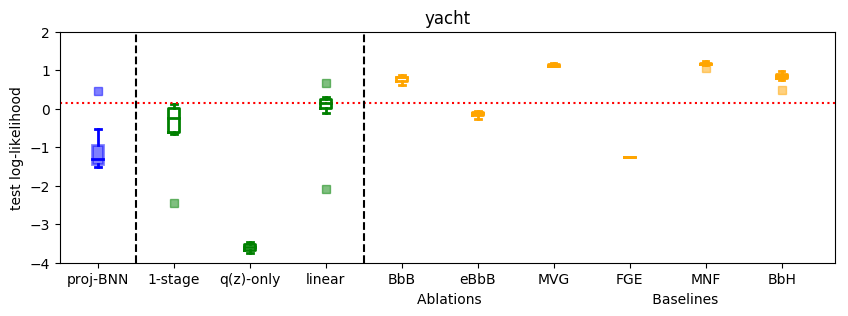}\includegraphics[width=0.5\textwidth]{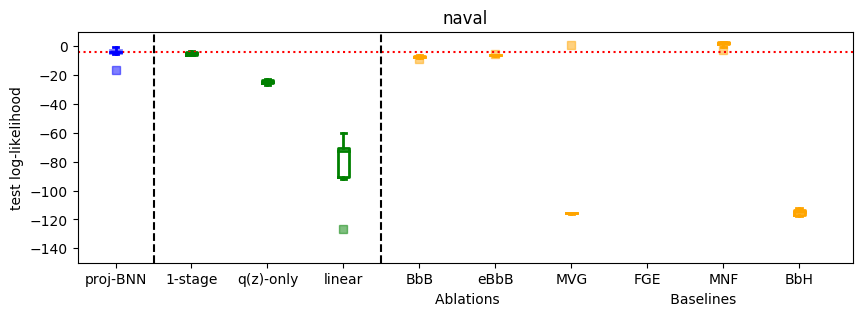}
	\caption{Extrapolation split. Metric $\mathbb{E}_{(\x^\star,\y^\star)} \left[\log p(\y_n^\star|\x_n^\star,\mathcal{D})\right]$.}
\end{figure*}
%%%

%%%   FIGURE
\begin{figure*}[h!]
	\includegraphics[width=0.5\textwidth]{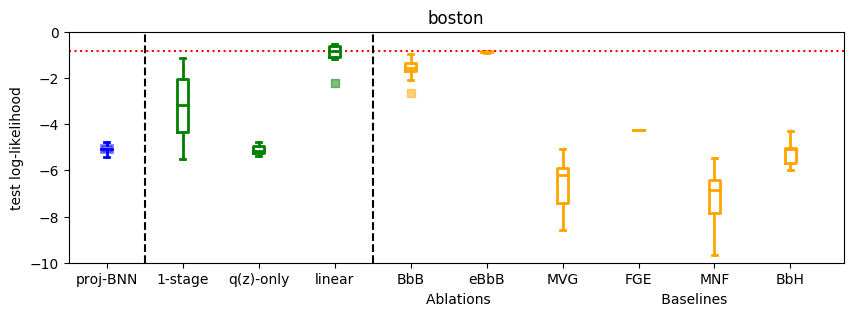}\includegraphics[width=0.5\textwidth]{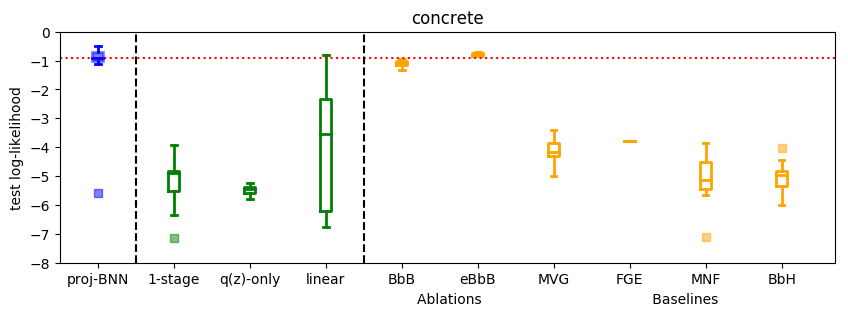}
	\includegraphics[width=0.5\textwidth]{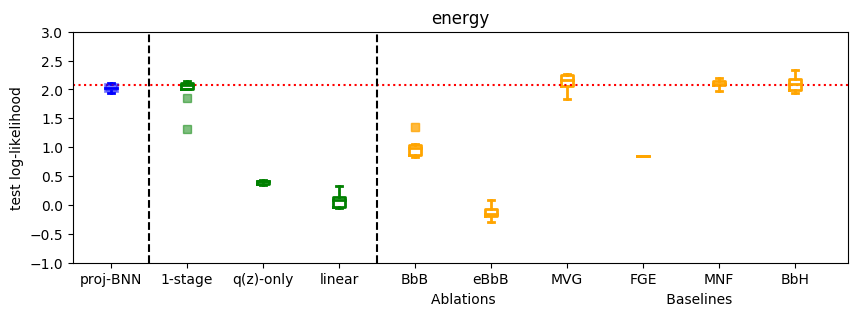}\includegraphics[width=0.5\textwidth]{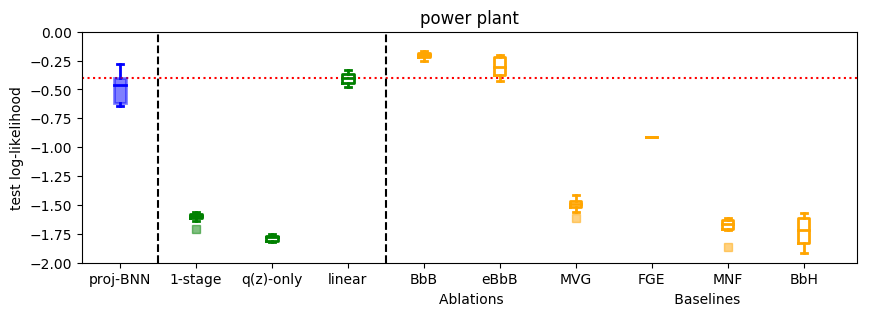}
	\includegraphics[width=0.5\textwidth]{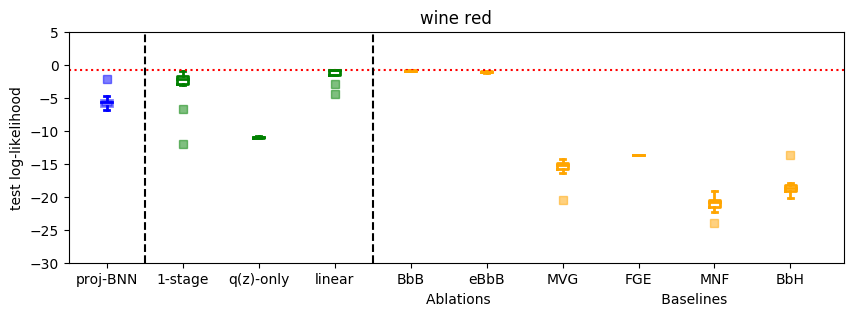}\includegraphics[width=0.5\textwidth]{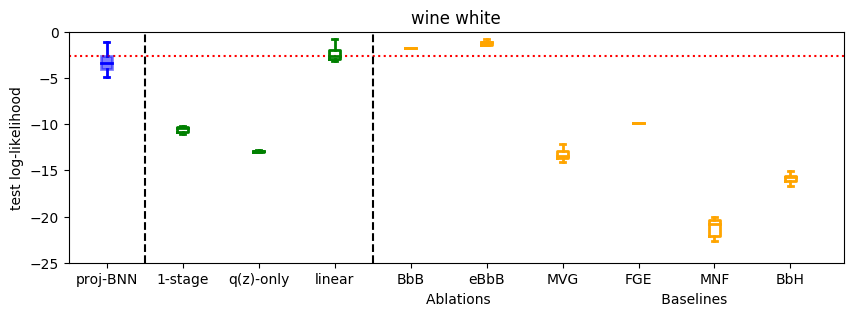}
	\includegraphics[width=0.5\textwidth]{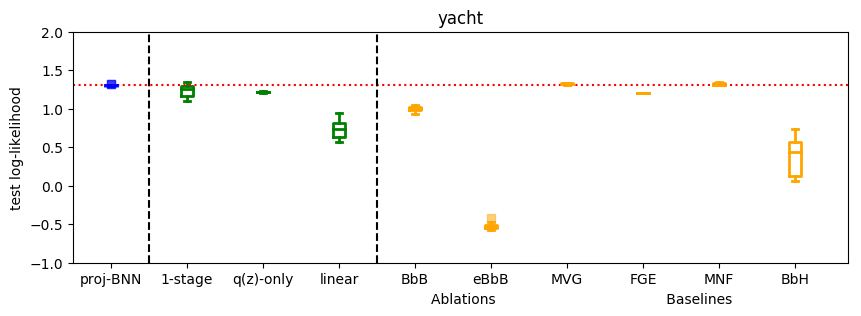}\includegraphics[width=0.5\textwidth]{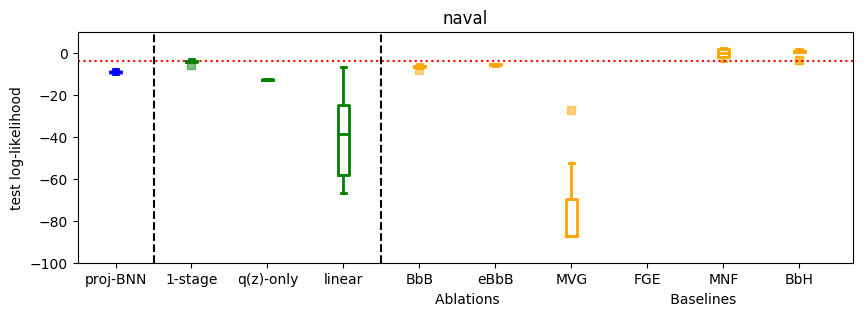}
	\caption{Interpolation split. Metric $\mathbb{E}_{(\x^\star,\y^\star)} \left[\log p(\y_n^\star|\x_n^\star,\mathcal{D})\right]$.}
\end{figure*}
%%%
%\subsection{Further 1-D regression toy examples}

%\subsection{Robustness across different initializations}

\subsection*{Proj-BNNs as broad priors over functions}

Proj-BNNs can be adapted to express broader priors over functions. In particular, consider a Proj-BNN variant where the auxiliary latent variable $\z$ is function specific. Given $M$ functions where each function provides us with a set of training instances, $\{\x_m, \y_m\}_{m=1}^M$, for each task $m$,  
\begin{eqnarray}
\z_m \sim  \mathcal{N}(\z_m\mid\mathbf{0}, \mathbf{I}), \, \W_m =  g_{\decp}(\z_m), \, \\
\y_m \sim \mathcal{N}(f_{\W_m}(\x_m), \sigma_y^2),
\label{eq:meta}
\end{eqnarray}
with the projection parameters $\decp \sim  p(\decp)$ shared across all functions. The function specific latent variables allow us to capture function specific variations while the share projection parameters allows for the sharing of statistical strength across functions. 

\begin{figure}
	\centering
	\begin{minipage}{.45\textwidth}
		\centering
		\includegraphics[width=1.0\textwidth]{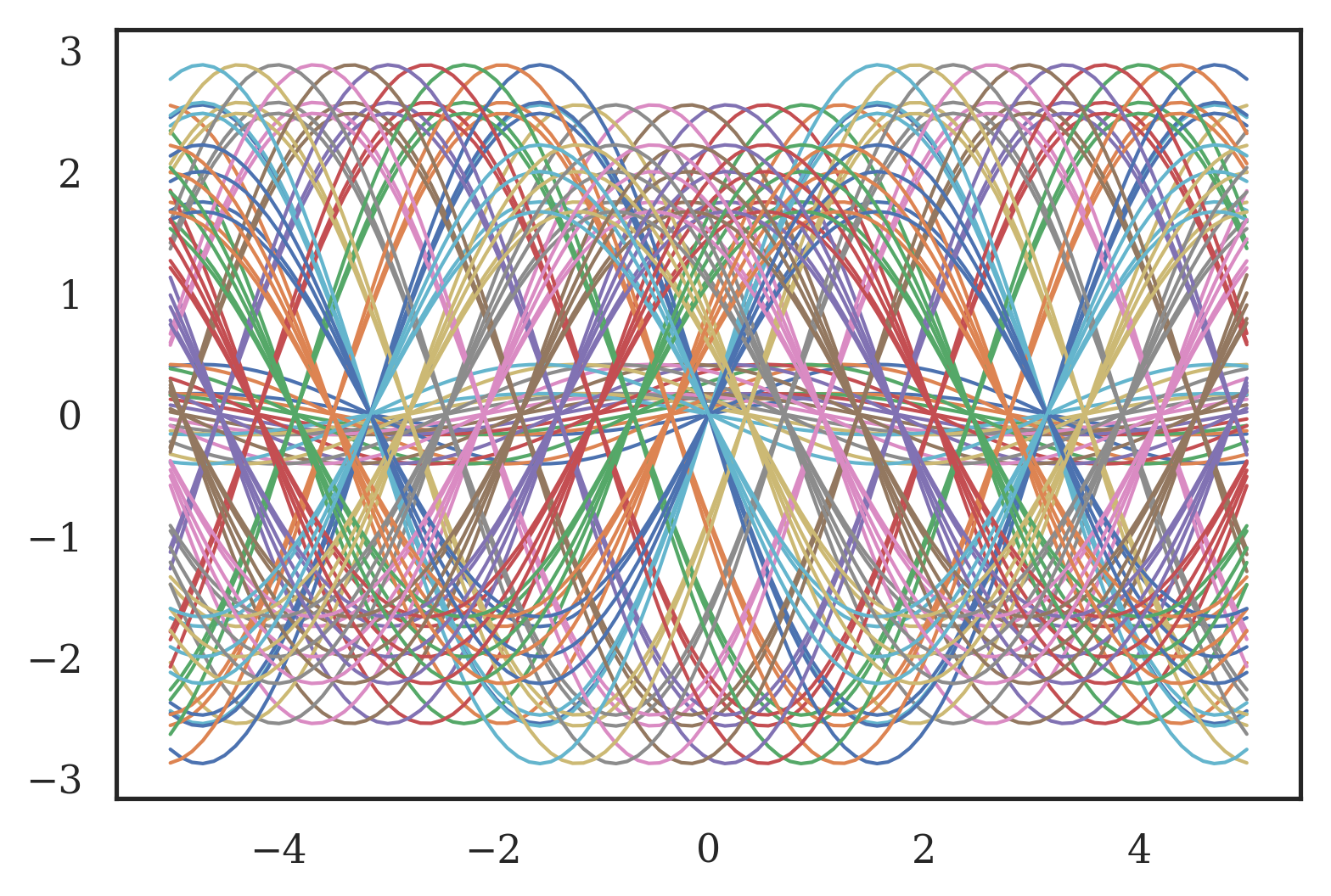}
		\caption{Visualization on training functions drawn from the family $y = a\times \text{sin}(x + b)$.}
		\label{fig:meta-train}
	\end{minipage}%
	\qquad \quad
	\begin{minipage}{.45\textwidth}
		\centering
		\includegraphics[width=1.0\linewidth]{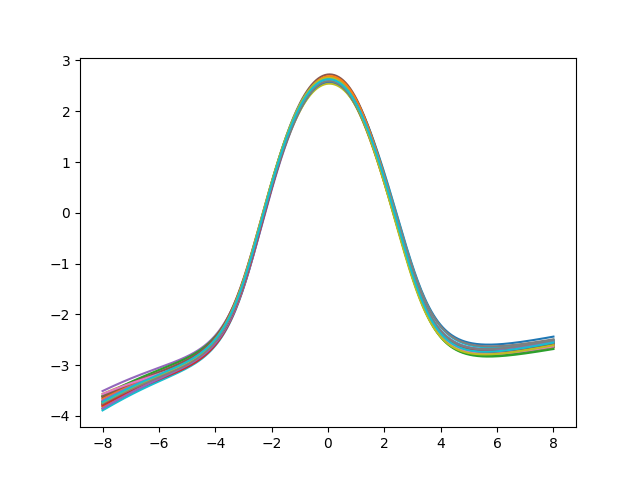}
		\captionof{figure}{Visualization of variance captured in $\phi$. Different colors correspond to different draws for a fixed $\z^s$ but varying samples of $\phi^s \sim q(\phi)$.}
		\label{fig:meta1}
	\end{minipage}
\end{figure}

\begin{figure*}
	\centering
	\includegraphics[width=0.95 \textwidth]{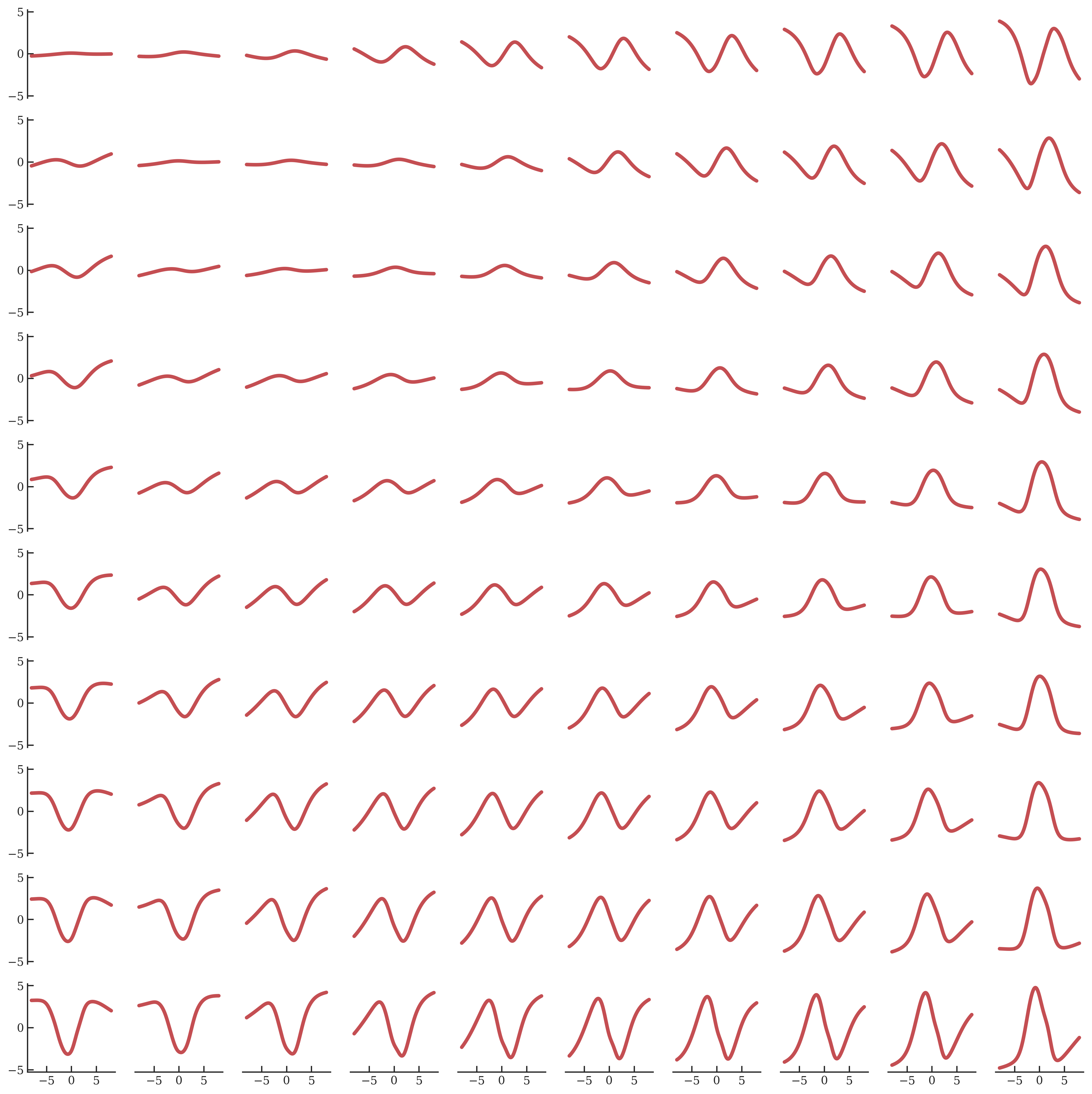}
	\caption{Visualization of the learned latent space $\z$. We explore the two dimensional space by transforming linearly spaced coordinates on the unit square through the inverse CDF of a standard Gaussian to produce instantiations of $\z$, $\z^s$. Using the variational mean $\tilde{\mu}_\phi$, we plot $y = f_{w^s}(x)$, where $w^s = g_{\tilde{\mu}_\phi}(\z^s)$.}
	\label{fig:meta}
\end{figure*}

To explore the effectiveness of such a model we created a toy 1-D regression dataset of sine functions, $y = a\times \text{sin}(x + b)$, of varying amplitudes and phases. We sampled $a \sim \text{Unif}(-3, 3)$ and varied $b$ by selecting equally spaced points in the range $[0, 2\pi]$. The set of training functions are shown in Figure~\ref{fig:meta-train}. Given the diversity of functions, a shared BNN struggles to learn meaningful mappings. In contrast, the Proj-BNN described in Equation~\ref{eq:meta} with $D_z=2$, and a $50$ unit network with tanh activations, learns an intuitive representation of the entire space (Figure~\ref{fig:meta}). Moreover, Proj-BNN decomposes uncertainty between $\z$ and $\phi$ (Figure~\ref{fig:meta1}). While $\z$ captures the structural differences between the functions, uncertainty in $\phi$ captures the uncertainty associated with a single function. These capabilities make Proj-BNN particularly attractive for multi-task and meta-learning applications, directions we plan to explore further in future work.

%Meta-learning, also known as “learning to learn”, is an increasingly popular field of research where models adapt to different environments or tasks rapidly with few training examples for each task~\citep{sutskever2014sequence,wang2016learning,vinyals2016matching,finn2017model}.
%
%The proposed approach proj-BNN can be applied for meta-learning scenarios by forcing the auxiliary latent variable $\z$ to be task-specific. In this Section, we show a proof-of-concept toy example where proj-BNN is able to learn a map of functions within the latent space $z$. We trained our model with a latent space $D_z=2$ on a dataset of $\mathrm{sin}$ functions with different amplitude, frequency, and sign function. Training a standard BNN with Bayes by Backprop~\cite{} using the whole dataset fails drastically. By constraining each task $m$ to have a different auxiliary latent variable $\z_m$, proj-BNN gives better performance in terms of marginal test log likelihood, and provides a probability distribution over function space that can be visualized by looking into the 
%%
%Figure~\ref{fig:meta} shows the predictive function means for different samples of $\z$. 

\end{document}

%% file: 00_prelude.tex
\def \c {\boldsymbol{c}}

\newcommand{\W}{\boldsymbol{w}}
\newcommand{\x}{\mathbf{x}}
\newcommand{\X}{\mathbf{X}}
\newcommand{\y}{\mathbf{y}}
\newcommand{\Y}{\mathbf{Y}}
\newcommand{\z}{\boldsymbol{z}}

\newcommand{\wc}{\mathbf{w_c}}
\newcommand{\wtc}{\hat{\mathbf{w}}_\mathbf{c}}

\newcommand{\vpphimu}{\boldsymbol{\tilde{\mu}}_\phi}
\newcommand{\vpphisig}{\boldsymbol{\tilde{\sigma}}_\phi}
\newcommand{\vpzmu}{\boldsymbol{\tilde{\mu}}_z}
\newcommand{\vpzsig}{\boldsymbol{\tilde{\sigma}}_z}

\newcommand{\E}[1]{ \ifthenelse{\equal{#1}{}} {\mathbb{E}} { \mathbb{E}_{#1} }}
\def \Eq {\mathbb{E}_q}

\newcommand{\encp}{\boldsymbol{\theta}}
\newcommand{\decp}{\boldsymbol{\phi}}
\newcommand{\vp}{\boldsymbol{\lambda}} % variational parameters

%% file: 01_intro.tex
\vspace{-0.9cm}
\section{Introduction}\label{sec:intro}

Deep learning provides a flexible framework for function approximation that has been adopted across many domains including machine vision, natural language processing, speech recognition, bioinformatics, and game-playing~\citep{lecun2015deep}.  Yet, deep models tend to overfit when the number of training examples is small~\citep{su2019one}; furthermore, their primary focus in practice is often on computing point estimates of model parameters, and thus, these models do not provide uncertainties for their predictions -- making them unsuitable for applications in critical domains such as personalized medicine.  Bayesian neural networks (BNN) promise to address these issues by modeling the uncertainty in the network weights, and correspondingly, the uncertainty in output predictions\citep{mackay1992practical,neal2012bayesian}.

Unfortunately, characterizing uncertainty over parameters of modern neural networks in a Bayesian setting is challenging due to the high dimensionality of the weight space and complex patterns of dependencies among the weights. In these cases, Markov-chain Monte Carlo (MCMC) techniques for performing inference often fail to mix across the weight space, and standard variational approaches not only struggle to escape local optima, but also fail to capture dependencies between the weights.

 A recent body of work has attempted to improve the quality of inference for Bayesian neural networks (BNNs) via improved approximate inference methods~\citep{graves2011practical, blundell2015weight, hernandez2016black}, or by improving the flexibility of the variational approximation for variational inference~\citep{louizos2017multiplicative}. In particular, variational families that explicitly model correlations between weights have been shown to be effective in approximate inference~\citep{sun2017learning, louizos2016structured, gal2016dropout}. However, many of these techniques only treat linear weight correlations while others face the challenge of performing inference over complex distributions over weights.

In this work, we introduce a novel inference approach in which we capture nonlinear latent structures amongst neural network parameters by learning a map of the weights onto a low-dimensional latent space.  Our approach takes advantage of the following insight: learning (standard network) \emph{parameters} is easier in the high-dimensional space, but characterizing (Bayesian) \emph{uncertainty} is easier in the low-dimensional space. Low-dimensional spaces are generally easier to explore, if we have fewer correlations between dimensions, and can be better captured by standard variational approximations (e.g. mean field).  At the same time, the non-linear transformation between latent space and weight space allows us to encode flexible approximating distributions for posteriors over the weights.  

Our main contribution is an inference framework that is able to exploit nonlinear latent structures in weight space.  We demonstrate, on synthetic datasets, the ability of our method to capture complex posterior distributions over weights by encoding them as distributions in latent space. We show that, as a result, our method is able to more accurately capture the uncertainty in the posterior predictive distribution. Finally, we demonstrate that, across a wide range of real-world data sets, our approach often produces more accurate predictions and higher quality predictive intervals compared to other state-of-the-art techniques.
% with highly compressed latent representations of the weights.

%% file: 02_literature_review.tex
 \section{Related Work}\label{sec:related} 
 
 % SG: Added some discussion about classic approaches. Could be a segue to a discussion on SG-MCMC methods
Classic work on Bayesian neural networks can be traced back to~\citep{WBuntine91, DMackay92, RNeal93}. \cite{RNeal93} introduced Hamiltonian Monte Carlo (HMC) to explore the Bayesian neural network posterior. \cite{DMackay92} and \cite{WBuntine91} instead relied on Laplace approximation to characterize the network posterior. While HMC remains the ``gold standard'' for posterior inference in BNNs, it does not scale well to modern architectures or large datasets. Similarly, vanilla application of the Laplace approximation has difficulty scaling to modern architectures with millions of parameters.
% MEL: Add SG-MCMC

\textbf{Structured Variational Approximations.}
Many recent works in variational inference have attempted to move beyond fully-factorized approximations of Bayesian neural network posteriors by modeling structured dependencies amongst BNN weights. Notably, matrix variate Gaussian distributions are proposed in \cite{louizos2016structured} to explicitly model the linear correlation between network weights. In a similar vein, our approach models non-linear dependencies and is most comparable to works that build flexible approximating families of weight distributions using auxiliary random variables, either through mixtures of simple distributions \citep{agakov2004auxiliary, maaloe2016auxiliary, ranganath2016hierarchical, salimans2013fixed} or via (non)-linear transformations of simple distributions~\citep{rezende2015variational, kingma2016improved, louizos2017multiplicative}.

\textbf{Normalizing Flows and Transformations.}
In particular,~\citet{rezende2015variational, kingma2016improved} apply normalizing flows to increase flexibility of the variational family distribution. In the context of BNNs, \cite{louizos2017multiplicative} assume normally distributed variational distributions for the weights with \emph{layer-wise} multiplicative noise that are \emph{linearly} projected onto a latent space via normalizing flows. However, they still assume that the posterior factorizes layer wise. In contrast, our approach learns \emph{non-linear} projections of the \emph{entire} network weights onto a latent space. Furthermore, our construction allows us to optimize a tighter bound on the log evidence.

A number of recent works use the idea of hypernetworks, neural networks that output parameters of other networks, to parametrize the variational distribution \citep{krueger2017bayesian, pawlowski2017implicit}. These works, however, do not model uncertainty in the latent projections. Moreover, they directly specify an implicit variational distribution over weights. Since, the implicit distribution's entropy is difficult to compute, authors have been forced to either use invertible projections thus trading off flexibility of the resulting distributions or otherwise approximate implicit weight densities potentially leading to severe local optima issues. In contrast, we incorporate uncertainty explicitly in both our generative and variational models and our approach of inference in the latent space avoids the challenges of dealing with implicit distributions.

In general, the vast majority of variational inference procedures for BNNs perform inference directly on the weight space \citep{sun2017learning, louizos2016structured, gal2016dropout} and face the challenge of working with complex distributions over high dimensional weights. In contrast, we perform inference in a lower dimensional latent space and thus solve an arguably easier problem. 

\textbf{Particle based inference}
Very recent work~\cite{wang2018function, sun2018functional}, building on ideas from Stein discrepancy~\cite{liu2016kernelized, liu2016stein}, perform inference directly in the function space. However, these approaches still maintain either a variational distribution over the weights~\cite{sun2018functional} or several particles in the weight space~\cite{wang2018function}. Our work is orthogonal, where we model the weights via a lower dimensional latent space, similar ideas may prove useful for inference in function space.

Stochastic gradient MCMC (SG-MCMC)~\cite{springenberg2016bayesian, li2016preconditioned} methods form the other popular class of BNN inference algorithms. While asymptotically exact, it is often difficult to determine when they have converged to their stationary distribution in the finite computation limit. Although, not our focus in the paper, SG-MCMC algorithms can be crafted for our projected BNN models.

\textbf{Weight embeddings/Modeling.}
In terms of modeling, our work is close in spirit to \citep{karaletsos_probabilistic_2018}, where the authors represent nodes in a neural network by latent variables via a \emph{deterministic} \emph{linear} projection, drawing network weights conditioned on those representations. In contrast to their approach, we learn a \emph{distribution} over \emph{non-linear} projections, and find a latent representation for the \textit{weights} directly rather than projecting the \textit{nodes}. % Up to our knowledge, this is the only other work where Bayesian neural network parameters are embedded into a low-dimensional space in the generative model. %We also propose a different multi-stage inference framework to enable learning a non-linear projection and latent representation, both with uncertainties.

Yet others~\cite{garnelo2018neural}, try to model flexible conditional distributions given arbitrary data instances. The resulting stochastic process termed Neural processes, are promising but may have difficulty scaling to even moderate dimensional data and have an orthogonal focus -- on fast adaptation and meta-learning. In contrast, our goal is to perform improved inference for BNNs. In the appendix, we sketch out a variant of our model that is closely related to neural processes.  
%Variational Implicit Processes (Ma et al., 2018) specify BNN priors and approximately fit a GP for the posterior,
% As you add in the neural process work, make sure that there's still a flow of ideas.  Maybe sub-headers for ensembles, bayesian projection/some kind of embedding types, non-bayesian dim red?

\textbf{Ensembles of Neural Networks.} Avoiding Bayesian inference completely, another line of work relies on (non-Bayesian) neural network ensembles to estimate predictive uncertainty~\citep{lakshminarayanan2017simple,pearce2018uncertainty}. The idea is simple: each network in the ensemble will learn similar values close to the training data, and different ones in regions of the space far from the training data. Whereas \cite{lakshminarayanan2017simple} rely on multiple random restarts and adversarial training, \cite{pearce2018uncertainty} introduce noise in the regularization term of each network, which directly relates to the randomized MAP sampling literature~\citep{lu_ensemble_2017, garipov2018loss}. In \citep{pearce2018uncertainty}, the authors derive a parallelism between their ensemble sampling approach and Bayesian behavior, but this only holds for infinite single-layer neural networks.
 % As you add in the neural process work, make sure that there's still a flow of ideas.  Maybe sub-headers for ensembles, bayesian projection/some kind of embedding types, non-bayesian dim red?
 % FDV: Can we make the variational statement more precise?  That is, why is drop-out related work?  Maybe we can lump all the non-distilled approaches together as ``Many approaches just generate particles, either toward the true posterior or some variational objective (CITE HMC/MCMC, DROPOUT, BOOTSTRAP).  Other work tries to use the samples to fit the final distribution.

\textbf{Compression of Neural Network Weights.}
Finally, the dimensionality reduction aspect of our modeling is comparable to neural network compression. There are a number of non-Bayesian methods for compressing standard neural networks, all of which, to our knowledge, rely on linear dimensionality reduction techniques on the space of weights and or network nodes \citep{denil_predicting_2013, sainath_low-rank_2013, xue_restructuring_2013, nakkiran_compressing_2015}. In contrast, our work is focused on non-linear dimensionality reduction in a Bayesian setting with the aim to improve uncertainty quantification.

%% file: 03_methods.tex
\section{Background and Notation}
\label{sec:background}

Let $\mathcal{D}=\{(\x_1,\y_1), \ldots (\x_N,\y_N) \}$ be a dataset of $N$ i.i.d observed input-output pairs.  We model this data by $\y =f_{\W}(\x) + \epsilon$, where $\epsilon$ is a noise variable and $\W$ refers to the weights of a neural network.\footnote{In this paper, $\W$ refers to both, network weights and biases, since each layer can be augmented with an extra input dimension containing 1's to account for the biases.}.  In the Bayesian setting, we assume some prior over the weights $\W \sim p(\W)$. One common choice is to posit i.i.d normal priors over each network weight $w_{i} \sim \mathcal{N} \left(0,\sigma^2_w \right)$.

Our objective is to infer the posterior distribution over functions $p(f_{\W}|\mathcal{D})$, which is equivalent to inferring the posterior distribution over the weights  $p(\W|\mathcal{D})$.  Given the posterior distribution, we model predictions for new observations and their associated uncertainties  through the posterior predictive distribution:
\begin{eqnarray}
	p(\y^{\star}|\x^{\star},\mathcal{D}) = \int p(\y^{\star}|\x^{\star},\W) p(\W | \mathcal{D}) d\W.
\end{eqnarray} 

The posterior $p(\W|\mathcal{D}) \propto p(\mathcal{D}|\W)p(\W)$ generally does not have an analytic form due to the non-linearity of $f_{\W}$; thus, one must resort to approximate inference techniques. Variational inference, for example, attempts to find a distribution $q_{\vp}(\W)$ that closely approximates the true posterior $p(\W|\mathcal{D})$. The measure of proximity is typically the KL-divergence:
\begin{align}
	D_{\mathrm{KL}} &\big(q_{\vp}(\W) || p(\W|\mathcal{D}) \big) \nonumber \\
	= & \; \int q_{\vp}(\W) [\log q_{\vp}(\W) - \log p(\W|\mathcal{D}) ] d\W \nonumber \\
	 = & -\mathcal{H}(q) - \Eq [\log p(\mathcal{D},\W) ] + \log p(\mathcal{D}) \nonumber \\
	 = & \; - \mathcal{L}(\vp) + \log p(\y|\x).\label{eq:KL}
\end{align}
where  $\mathcal{L}(\vp) = \mathcal{H}(q) + \Eq [\log p(\mathcal{D},\W) ]$ is the evidence lower bound (ELBO) on the marginal likelihood $\log p(\y|\x)$. Approximate inference in this case can be cast as the problem of optimizing $\mathcal{L}(\vp)$ with respect to $\vp$, since minimizing the KL-divergence between $q_{\vp}(\W)$ and $p(\W|\mathcal{D})$ is equivalent to maximixing the lower bound $\mathcal{L}(\vp)$, as shown in Eq.~\eqref{eq:KL}.

For BNNs, the posterior distribution $p(\W | \mathcal{D})$ often contains strong correlations, such that the optimization of $\vp$ is prone to get stuck in local optima.  Moreover, simple variational families $q_{\vp}(\W)$, such as mean field approximations, fail to capture those correlations.

\section{Latent Projection BNN}
\label{sec:methods}
\subsection{Generative Model}
In our approach, which we call \emph{Projected Bayesian Neural Network (Proj-BNN)}, we posit that the neural network weights $\W$ are generated from a latent space or \textit{manifold} of much smaller dimensionality. That is, we assume the following generative model:
\begin{eqnarray}
\z \sim  p(\z), \,  \decp \sim  p(\decp), \, \W =  g_{\decp}(\z), \, \y \sim \mathcal{N}(f_{\W}(\x), \sigma_y^2)
\end{eqnarray}
where  $\W$ lies in $\mathbb{R}^{D_{w}}$, the latent representation $\z$ lie in a lower dimensional space $\mathbb{R}^{D_{z}}$, and $\decp$ parametrizes the arbitrary projection function $g_{\decp}:\mathbb{R}^{D_z} \to \mathbb{R}^{D_w}$. Given this generative model, our objective is to compute the joint posterior distribution $p(\z,\decp |\y,\x)$ over both the latent representation $\z$ and the latent projection parameters ${\decp}$.

% FDV: Sec. 4 - Is this how MVG is laid out?  Don't have the paper with me.  (I liked your idea of keeping that description parallel)

% FDV: Sec. 4.2, should we note that since mean-field on phi, it will be around a mode -> we find a reasonable starting mode?  we kind of say it, but could be more explicit
\subsection{Inference}
Following the same structure in the generative model, we propose a variational distribution $q_{\vp}(\z,\decp) = q_{\vp_{z}}(\z) q_{\vp_{\phi}}(\decp)$ such that:
\begin{align}
\z  \sim  q_{\vp_{z}}(\z), \;\decp  \sim  q_{\vp_{\phi}}(\decp), \; \W  =  g_{\decp}(\z).
\end{align}
In particular, we posit a mean-field posterior approximation for each independent term $q_{\vp_{z}}(\z)  \doteq \mathcal{N}(\vpzmu,\vpzsig \mathbf{I})$ and $q_{\vp_{\phi}}(\decp)  \doteq \mathcal{N}(\vpphimu,\vpphisig \mathbf{I})$, 
where $\vpzmu,\vpphimu$ and $\vpphisig,\vpphisig$ refer to the mean and standard deviation vectors of each Normal variational distribution respectively, $\vp_{z} = \{ \vpzmu,\vpzsig \}$, and $\vp_{\phi} = \{ \vpphimu,\vpphisig \}$. Note that, although we adopt a fully factorized posterior approximation for $\z$ and $\decp$, the induced posterior approximation $q_{\vp}(\W)$ over the network weights $\W$ can capture complex dependencies due to the non-linear transformation $g_{\decp}$.

Given such variational distribution, a straightforward inference algorithm is to use black-box variational inference (BBVI)~\citep{ranganath2014black} with the reparametrization trick~\citep{kingma2015variational} to minimize the joint evidence lower bound (ELBO) $\mathcal{L}(\vp)$ given by:
\begin{align}
\mathcal{L} & (\vp) = \Eq \Big[\log  p \big(\y|\x,g_{\decp}(\z) \big)\Big] \nonumber \\
& - D_{\mathrm{KL}} \big(q_{\vp_z}(\z)||p(\z) \big)  - D_{\mathrm{KL}} \big(q_{\vp_\phi}(\decp)||p(\decp) \big). \label{eq:ELBO2}
\end{align}
However, jointly optimizing the projection parameters $\decp$ and latent representation $\z$ is often a hard optimization problem; direct optimization of the ELBO in Eq.~\eqref{eq:ELBO2} is prone to local minima and might lead to poor performance solutions. To alleviate this issue, we additionally come up with an intelligent initialization (no uncertainty) for $\decp$, and then perform variational inference using BBVI. In the following, we describe the complete inference framework in three stages.

\textbf{Characterize the space of plausible weights.}
First, we seek to gather a diverse set of weight parameters (without considering uncertainty), which will be used to learn a smart initialization of $\decp$ afterwards. As explained in Section~\ref{sec:intro}, learning parameters in the high-dimensional $\W$-space is easy, the difficult part is to get accurate uncertainty estimations.
 More precisely, we collect multiple, high-quality candidate weight solutions $\{ \wc^{(r)} \}^{R}_{r=1}$ by training a non-Bayesian neural network and sampling around the MAP solution, using Fast Geometric Ensembling (FGE)~\citep{garipov2018loss,izmailov2018averaging}. This is equivalent to training an ensemble of neural networks from multiple restarts, but in the time required to train a single model, and avoiding identifiability problems over the weight space.\footnote{Solutions corresponding to weight permutations are naturally avoided by sampling close to the same local optimum.} %\footnote{Neural network ensembles have been shown to provide diverse sets of predictive functions while remaining accurate~\cite{lakshminarayanan2017simple, pearce2018uncertainty}.}
 FGE allow us to recover a variety of weights which will lead to similar function values where there is training data, but different outputs otherwise.
 Although there exist methods in the literature to force diversity in the solutions of neural network ensembles, we opt for FGE for simplicity and computational efficiency.
%

%\vspace{-0.3cm}
\textbf{Learn a point-estimate for the projection function.}
In order to find an intelligent initialization in the BBVI algorithm for the projection function $g_{\decp}$, we perform dimensionality reduction on the previously collected sets of weights $\{ \wc^{(r)} \}^{R}_{r=1}$. We opt for an autoencoder to account for non-linear complex transformations, but other alternatives can also be applied. Let $f_{\encp}: D_w \rightarrow D_z$ and $g_{\decp}: D_z \rightarrow D_w$ denote the encoder and decoder, respectively, of the autoencoder $h_{\encp,\decp}=g_{\decp} \circ f_{\encp}$. Our aim is to find a point estimate for the parameters of the projection.

While we want to find latent projections that minimize the reconstruction error of the weights, at the same time, we also need to explicitly encourage for projections that will map into weights that yield high log likelihood values for our original training data $\mathcal{D}$. We find that, in practice, the explicit constraint on the predictive accuracy of reconstructed weights $\wtc^{(r)} = h_{\encp,\decp}\left(\wc^{(r)}\right)$ is required since weights that are ``similar'' in Euclidean norm may yield models of very different predictive qualities.\footnote{If we only optimize for minimum weight reconstruction, we find projections whose reconstructed weights have lower quality in terms of test log likelihood $p(\y|\x,\wtc^{(r)})$.} This results in the following loss to minimize:
\begin{align}
 \mathcal{L}(& \encp,\decp) = \frac{1}{R} \sum^R_{r=1} \left(\wc^{(r)}  - h_{\encp,\decp}\left(\wc^{(r)}\right) + \gamma^{(r)} \right)^2 \nonumber \\
	 + & \; \beta \; \mathbb{E}_{(x,y) \sim \mathcal{D} } \left[ \frac{1}{R} \sum^R_{r=1} \log p\left(y|x, h_{\encp,\decp}\left(\wc^{(r)}\right) \right)\right], \label{eq:pcAE}
\end{align}
where $\gamma^{(r)} \sim \mathcal{N}(0,1)$ is an additional input noise term that makes training more robust. The first term in Eq.~\eqref{eq:pcAE} corresponds to the average mean square error of each $\wc^{(r)}$ and its reconstructed version $\wtc^{(r)}$, while the second term accounts for the reconstruction error (in terms of log likelihood) of the original output data $\y$ given $\x$ and $\wtc^{(r)}$. We call this approach a \emph{prediction-constrained} autoencoder. Prediction-constrained models have previously been introduced in the context of mixture and topic models~\citep{hughes2017prediction,hughes2018semi}.

\textbf{Learn the approximate posterior $q_{\vp}(\z,\decp).$}
Given the point-estimate projection parameters ${\decp^{\star}}$ from the previous stage, we can now initialize the mean variational parameters for $q_{\vp}(\z,\decp)$, and perform principled posterior approximation using black-box variational inference. For simplicity, we assume a mean-field structure $q_{\vp}(\z,\decp) = q_{\vp_{z}}(\z) q_{\vp_{\phi}}(\decp)$ for the latent representation $\z$ and projection parameters $\decp$.
We optimize the variational parameters $\vp = \{ \vp_{z}, \, \vp_{\phi}\}$ to minimize the KL-divergence $D_{\textrm{KL}} \big(q_{\vp}(\z,\decp)||p(\z,\decp|\y,\x) \big)$, which is equivalent to maximizing the evidence lower bound (ELBO) from Eq.~\eqref{eq:ELBO2}.
% {eq:ELBO2}
%To facilitate the optimization task, we first optimize the variational distribution in latent space $q_{\vp_{z}}(\z)$ (assuming $\decp$ fixed) via black-box variational inference (BBVI)~\citep{ranganath2014black} with the reparametrization trick~\citep{kingma2015variational}, after which we jointly fine-tune the uncertainty in both the latent space and the projection parameter. In other words, we proceed to optimize two different evidence lower bounds (ELBO). We first assume an approximate mean-field variational distribution $q_{\vp_{z}}(\z)$ in the latent space. The following lower-bound $\mathcal{L} (\vp_{z})$ on the marginal log-likelihood can be derived:
%\begin{eqnarray}
%\log p(\y|\x) \geq
%\int q_{\vp_{z}}(\z) \log \frac{p \big(\y|\x,g_{\decp}(\z) \big)p(\z)}{q_{\vp_z}(\z)}d\z \\
%\mathcal{L} (\vp_{z}) = \Eq \Big[\log p \big(\y|\x,g_{\decp}(\z) \big)\Big] - D_{\mathrm{KL}} \big(q_{\vp_z}(\z)||p(\z) \big). \label{eq:ELBOs}
%\end{eqnarray}
The expectation in Eq.~\eqref{eq:ELBO2} and gradient can be estimated using simple Monte Carlo integration along with the reparametrization trick~\citep{kingma_auto-encoding_2013,rezende_stochastic_2014}:
\begin{eqnarray}
\Eq \Big[\log p \big(\y|\x,g_{\decp}(\z) \big)\Big] \approx \frac{1}{S}\sum^S_{s=1} \log p \big(\y|\x,g_{\decp^{(s)}}(\z^{(s)}) \big),\nonumber
\end{eqnarray}
where $\z^{(s)} = \boldsymbol{\epsilon}^{(s)}_z\; \vpzsig + \vpzmu$, $\vp_{z}= \{ \vpzmu,\vpzsig\}$ are the variational parameters of the distribution $q_{\vp_{z}}(\z)$, $\decp^{(s)} = \boldsymbol{\epsilon}^{(s)}_\phi\; \vpphisig + \vpphimu$, $\vp_{\phi}= \{ \vpphimu,\vpphisig\}$ are the variational parameters of the distribution $q_{\vp_{\phi}}(\decp)$,  and $\boldsymbol{\epsilon}^{(s)}_z,\,\boldsymbol{\epsilon}^{(s)}_\phi \sim  \mathcal{N}(0,\mathrm{I})$ are the auxiliary noise variables for the reparametrization trick. 
The KL-divergence terms in Eq.~\eqref{eq:ELBO2} can be computed in closed form, as all distributions $p(\z)$, $p(\decp)$, $q_{\vp_z}(\z)$, and $q_{\vp_{\decp}}(\decp)$ are normal-distributed.

 %Finally, we proceed to optimize all the variational parameters $\vp$ jointly by minimizing the augmented ELBO $\mathcal{L}(\vp)$ from Eq.~\eqref{eq:ELBO2}. 
 
 The variational parameters $\vpphimu$ are initialized to the point-estimate $\decp^{\star}$ computed in the previous stage, whereas the variational parameters $\vp_z$ and $\vpphisig$ are initialized randomly.\footnote{In practice, initializing $\vpphisig$ to small values gave better performance, i.e., $\log \vpphisig \sim \mathcal{N}(-9,0.1)$.} Algorithm~\ref{algo:framework} summarizes the three-step framework for tractable inference.

%%%   ALGO
\begin{algorithm}
	%\SetAlgoLined
	\textbf{Input:} observations $\mathcal{D}=\{(\x_n,\y_n)\}^N_{n=1}$
	%initialization\;
	\begin{enumerate}
		\item Gather multiple sets of weights $\{ \wc^{(r)} \}^{R}_{r=1}$ via Fast Geometric Ensembling~\citep{garipov2018loss}.
		\item Train a prediction-constrained autoencoder $h_{\encp,\decp}$ using $\{ \wc^{(r)} \}^{R}_{r=1}$ as input data to minimize the loss function described in Eq.~\eqref{eq:pcAE}.
		\item Perform BBVI to learn an approximate posterior distribution over latent representations $\z$ and projection parameters $\decp$.
		\begin{itemize}
			%\item Optimize ELBO $\mathcal{L}(\vp_z)$ in Eq.~\eqref{eq:ELBOs} to obtain $q_{{\vp}_{z}}(\z)$ closest to $p(\z|\y,\x,\decp)$ in terms of KL-divergence.
			\item Initialize mean variational parameters $\vpphimu$ for $\decp$ with the autoencoder solution.
			\item Optimize ELBO $\mathcal{L}(\vp)$ in Eq.~\eqref{eq:ELBO2} to obtain $q_{{\vp}}(\z,\decp)$ closest to $p(\z,\decp|\y,\x)$ in terms of KL-divergence.
		\end{itemize}
	\end{enumerate}
	\textbf{Result:} Approximate posterior $q_{\vp}(\z,\decp)$
	\caption{Inference for Proj-BNN}
	\label{algo:framework}
\end{algorithm}

%% file: 04_results.tex
 \section{Results}\label{sec:results}
This section contains results on synthetic and real-world datasets to illustrate the performance and potentials of the proposed approach (Proj-BNN). We show that Proj-BNN provides flexible posterior approximations, resulting in better uncertainty estimations, and improved model generalization in terms of held-out test log likelihood across multiple datasets.

We compare Proj-BNN to the following baselines: Bayes by Back Prop (BbB)~\citep{blundell2015weight}, Multiplicative Normalizing Flow (MNF)~\citep{louizos2017multiplicative}, Matrix Variate Gaussian Posteriors (MVG)~\citep{louizos2016structured}, Bayesian Hypernetworks (BbH)~\citep{pawlowski2017implicit}, and Fast Geometric Ensembling (FGE)~\citep{garipov2018loss}.

% FDV: fig. 1 -> x-axes!! not fair to be different! all text too tiny.
% FDV: combine fig. 2 and 3 -> about same thing

\subsection{Synthetic Data}
First, we demonstrate on synthetic data that, by performing approximate inference on $\z$, we can better capture uncertainty in the posterior predictive. Furthermore, we show that when the space of plausible weights for a regression model is complex, capturing this geometry in approximate inference is difficult; in contrast, performing inference in a simpler latent space allows for better exploration of the solution set in the weight space of the regression model.

% FDV: The x-ranges on the figures are different -> can be misleading because then we focus on extrapolation for some algorithms and not for others.  Also, I thought we had an HMC baseline for this small example?  We've spent so much time poking at papers that don't have this baseline, let's not fall into the same trap :)

 %%%   FIGURE
\begin{figure*}[h!]
	\centering
	\subfloat[][Proj-BNN ($D_z=2$)]{\includegraphics[width=0.25\textwidth, height=39mm]{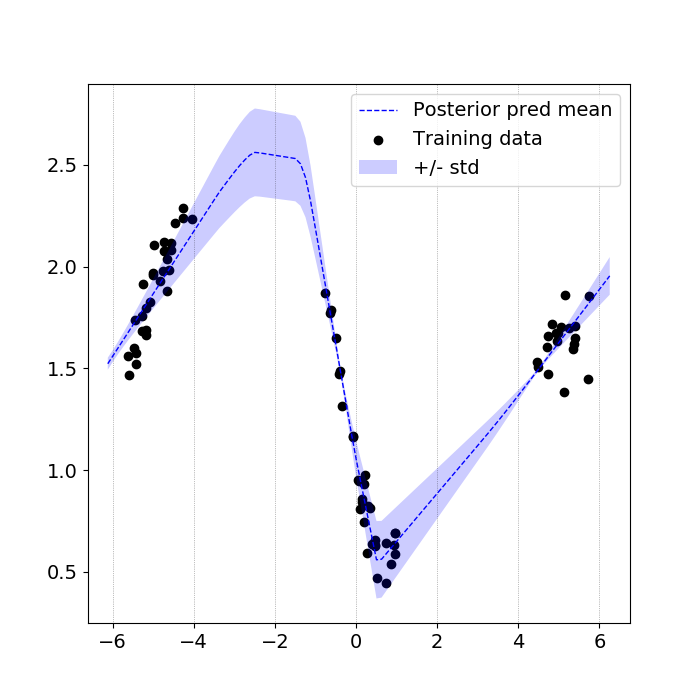}}
	%\subfloat[][Proj-BNN ($D_z=100$)]{\includegraphics[width=0.25\textwidth,  height=39mm]{./figs/toy_uncertainties/LP-BNN-dim_u=100.png}}
	\subfloat[][BbB]{\includegraphics[width=0.25\textwidth,  height=39mm]{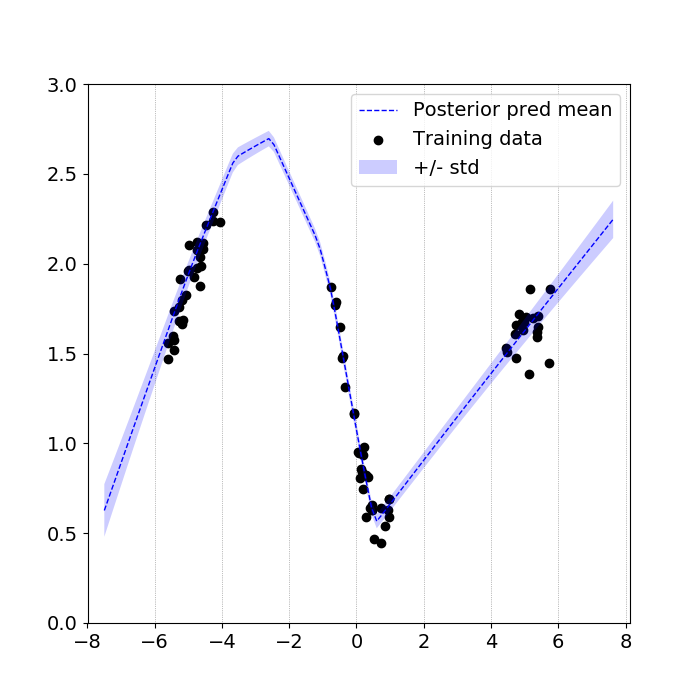}}
	\subfloat[][MNF]{\includegraphics[width=0.25\textwidth,  height=39mm]{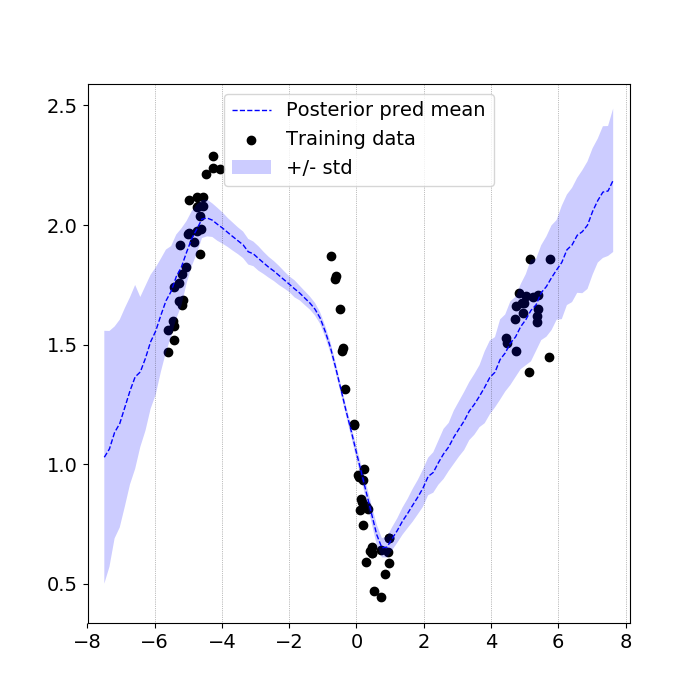}}\subfloat[][MVG]{\includegraphics[width=0.25\textwidth,  height=39mm]{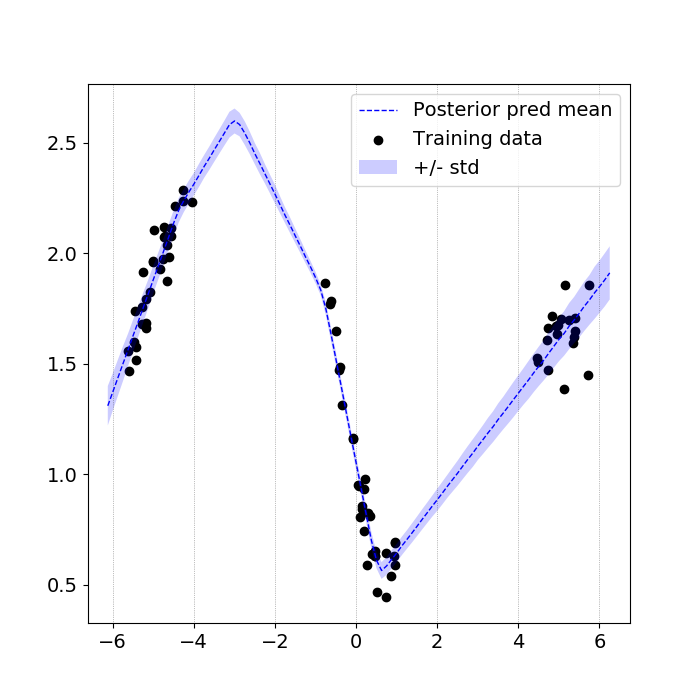}}
	\caption{Inferred predictive posterior distribution for a toy data set drawn from a NN with 1-hidden layer, 20 hidden nodes and RBF activation functions.  \textbf{Proj-BNN is able to learn a plausible predictive mean and better capture predictive uncertainties.} 
		\label{fig:toy_data1_post_pred}
	}
\end{figure*}
%%%

%\paragraph{Inference in latent space can provide better estimates of posterior predictive uncertainty.}
\textbf{Better estimates of uncertainty.}
Figure \ref{fig:toy_data1_post_pred} compares the posterior predictive distributions obtained by Proj-BNN against BbB, MNF, and MVG. The data is generated by sampling points non-uniformly from a function represented by a feedforward network (with 1 hidden layer, 20 hidden nodes and RBF activation centered at 0 with length scale 1) whose weights are obtained by applying a fixed linear transform to a fixed latent vector $\z$. We observe that our method is able to obtain a mean posterior predictive that fits the data and is furthermore able to capture more uncertainty in the posterior predictive. Notably, baseline methods tend to underestimate predictive uncertainty, especially in places with few observations, and thus produces over-confident predictions.

%%%   FIGURE
\begin{figure*}[t]
	\centering
	\subfloat[][Projection of true weights\label{fig:2a}]{\includegraphics[width=0.31\textwidth, height=32mm]{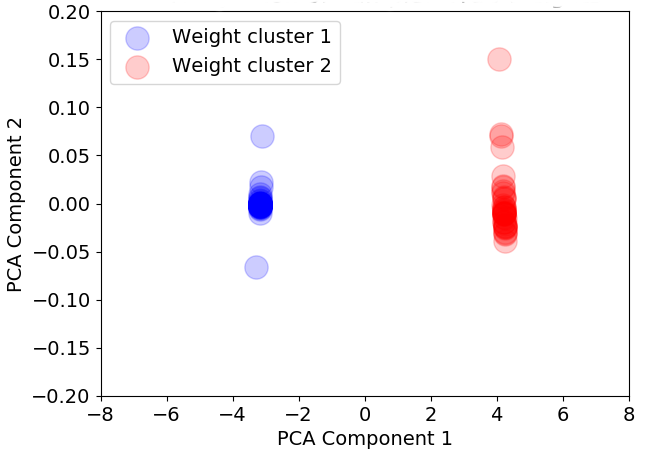}}\hspace{0.1cm}
	\subfloat[][BbB posterior over weights]{\includegraphics[width=0.31\textwidth, height=32mm]{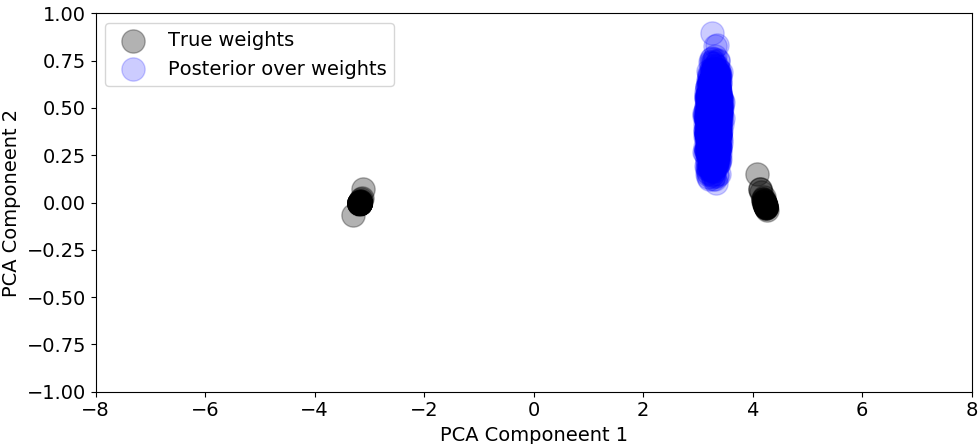}}\hspace{0.1cm}
	\subfloat[][Proj-BNN posterior over weights]{\includegraphics[width=0.31\textwidth, height=32mm]{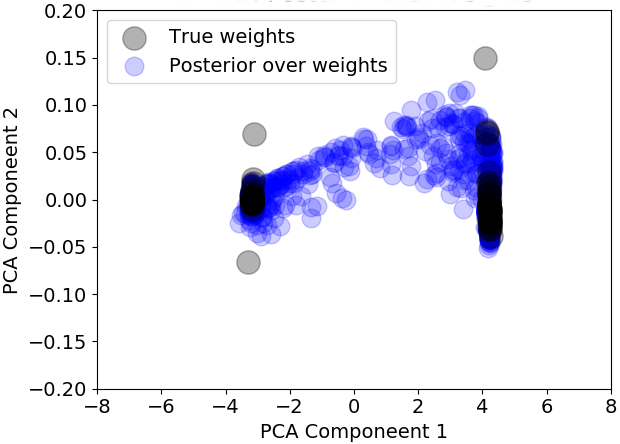}}
	
	\subfloat[][Functions from true weights]{\includegraphics[width=0.31\textwidth, height=32mm]{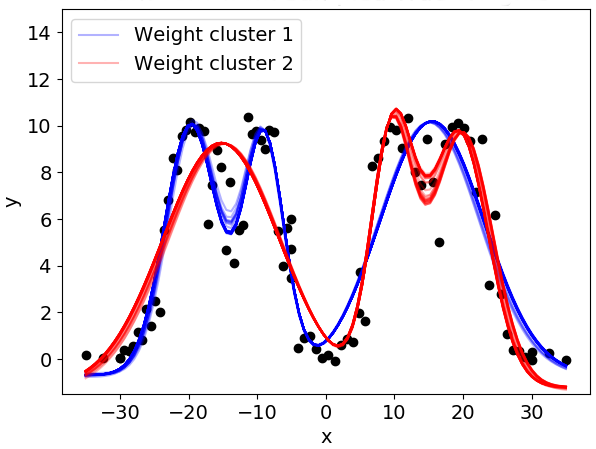}}\hspace{0.1cm}
	\subfloat[][BbB posterior predictive]{\includegraphics[width=0.31\textwidth, height=32mm]{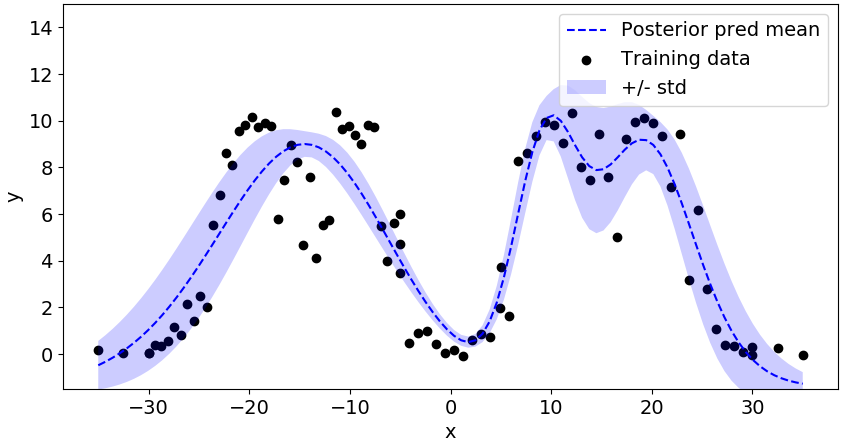}}
	\subfloat[][Proj-BNN posterior predictive]{\includegraphics[width=0.31\textwidth, height=32mm]{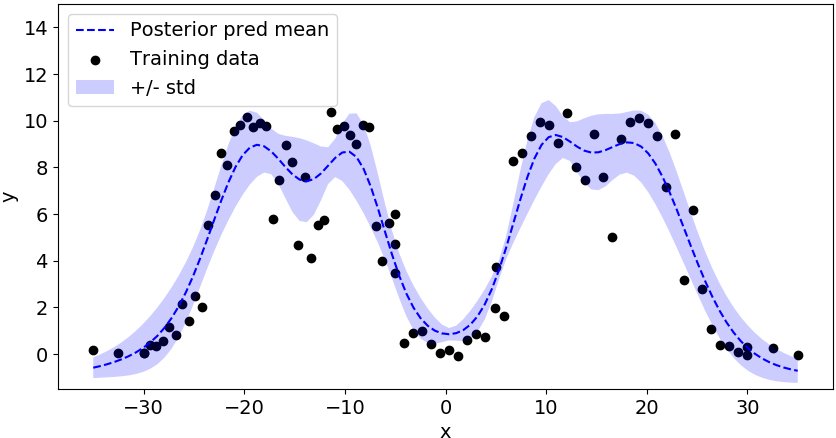}}\hspace{0.1cm}
	\caption{
		\textbf{Learning a variational posterior over $\z$ captures both modes in the weight space.} 
		 (a) PCA projection of sampled true weights of the feedforward network. (b) Variational posterior over weights learned by performing inference directly on $\W$, using Bayes by Back Prop (BbB); only one mode captured. (c) Variational posterior over weights, $\W$, obtained by transforming the variational posterior over $\z$. (d) Examples of functions corresponding to weights sampled from each weight cluster. (e) Posterior predictive using BBB; only three modes captured. (f) Posterior predictive using Proj-BNN; all four modes captured.}
	\label{fig:toy_data_posterior_weights}
\end{figure*}
%%%

%\paragraph{Inference in latent space can improve posterior predictive quality by capturing complex geometries of the weight posterior.}
\textbf{Capturing complex geometries in $\W$-space.}
We argue that the reason for the observed improvement in the quality of posterior predictive obtained by our method is often due to the fact that it is difficult to approximate complex geometries of the solution set in the weight space of sophisticated regression models. Mapping the solution set onto a simpler region in a lower dimensional latent space allows for more efficient approximations. In Figure~\ref{fig:2a}, we visualize samples of plausible weights for a feedforward network (with a single hidden layer, \emph{three} nodes and RBF activation function) fitted to a data set with \emph{four} modes.\footnote{The true sampled weights are obtained by fitting the data multiple times from random weight initializations.} We see that the solution set in the weight space for this model is naturally bimodal, where each mode corresponds to functions that fit a particular choice of three of the four modes in the data.

Figure~\ref{fig:toy_data_posterior_weights} shows that direct variational approximation of the posterior over weights is only able to capture one of the modes in the solution set, while approximating the posterior over $\z$'s using our method captures both modes. For the latter, we trained a decoder $g_{\decp}(\cdot)$ that maps an isotropic 2-D Gaussian to weights that we sampled from the solution set. As a result, the posterior predictive mean obtained by our method is able to approximate the four modes, while the predictive mean obtained from directly approximating the posterior over $\W$ can approximate only three of the four modes.

\subsection{Real Data}

\begin{figure*}[h!]
	\includegraphics[width=0.5\textwidth]{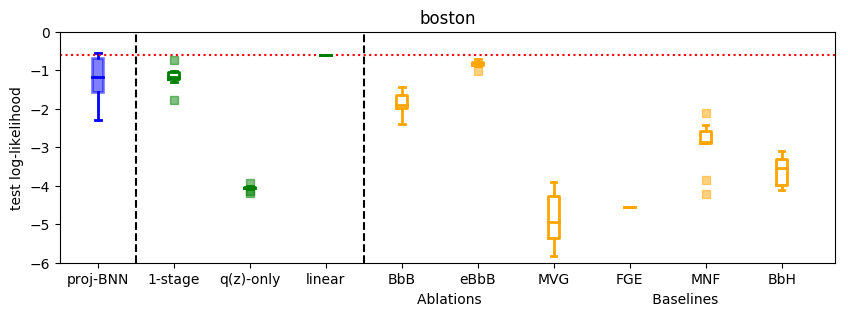}\includegraphics[width=0.5\textwidth]{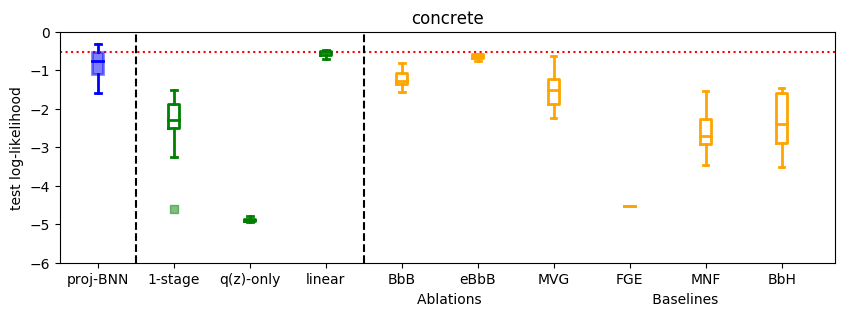}
	\includegraphics[width=0.5\textwidth]{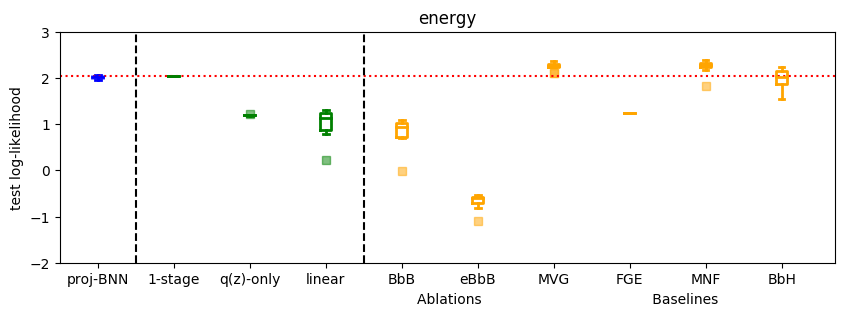}\includegraphics[width=0.5\textwidth]{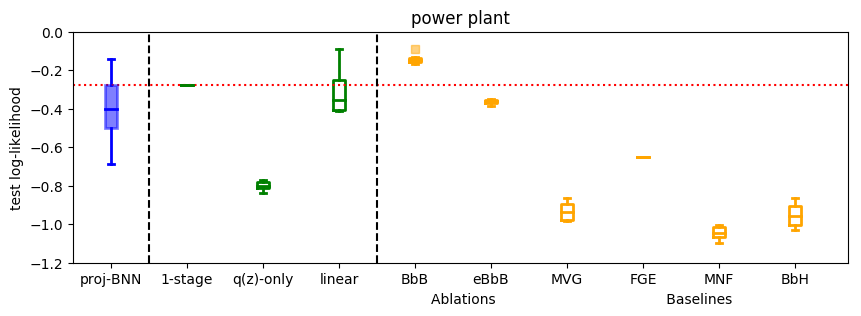}
	\includegraphics[width=0.5\textwidth]{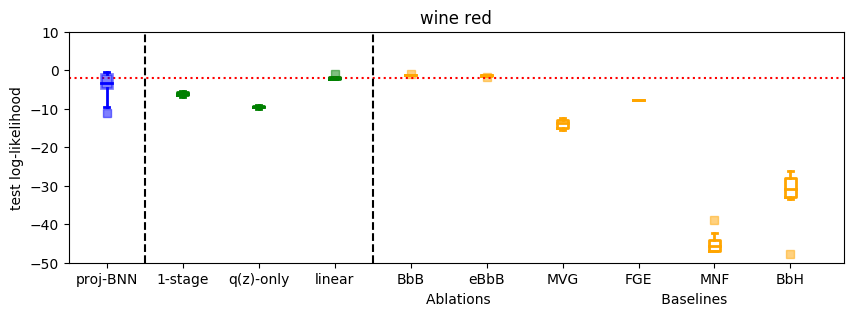}\includegraphics[width=0.5\textwidth]{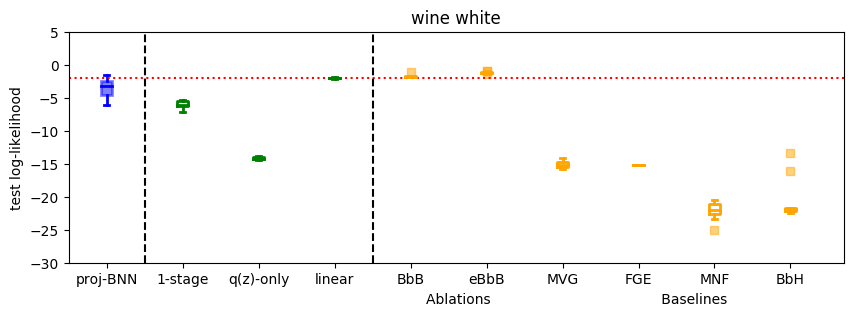}
	\includegraphics[width=0.5\textwidth]{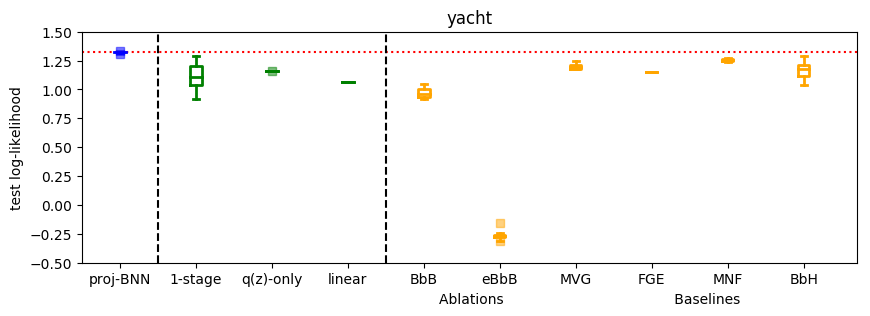}\includegraphics[width=0.5\textwidth]{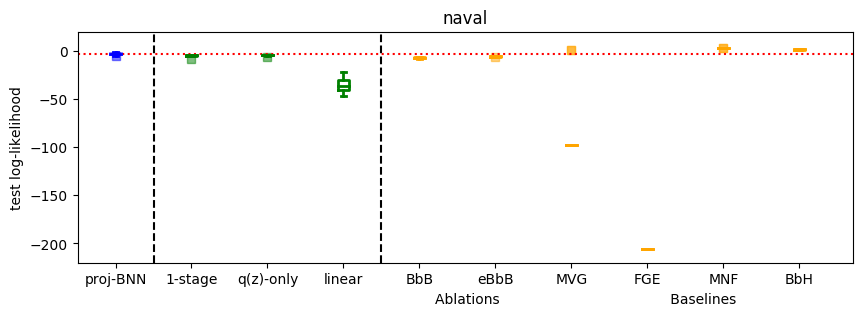}
		\caption{\textbf{Test log-likelihood for UCI benchmark datasets for best dimensionality of $z$-space.} Red dotted horizontal line corresponds to Proj-BNN performance (our approach) or one of its variants.
		Baselines methods: 1) BbB: mean field (Blundell, et.al 2015); 2) eBbB: extended BbB, with 1 layer of 3000 nodes; 3) MVG: multivariate Gaussian prior BNN (Louizos et.al, 2016); 4) FGE Fast Geometric Ensembling (Gastarov et.al, 2018); 5) MNF: multiplicative normalizing flow (Louizos et.al, 2017); 6) BbH: Bayesian hypernetworks (Pawlowski et.al, 2017).
		Ablations of Proj-BNN are: Proj-BNN with linear projections (linear), Proj-BNN without training the autoencoder, i.e., only stage 3 in Alg.~\ref{algo:framework} (1-stage), Proj-BNN modeling uncertainty only in $\z$ ($q(\z)$-only).}
		%\textbf{In all but two cases Proj-BNN performs better or as well as the baselines.}}
	\label{fig:UCI_testll}
	%\caption{Random Split. Metric $\mathbb{E}_{(\x^\star,\y^\star)} \left[\log p(\y_n^\star|\x_n^\star,\mathcal{D})\right]$}
\end{figure*}

\textbf{Simulation settings.}
We perform nonlinear regression on eight UCI datasets, listed in Figure \ref{fig:UCI_testll}. In all the following experiments, we use a random train-test-validation split of 80-10-10. All datasets are normalized in a preprocessing step to have zero mean and unit standard deviation.
For all inference methods, We train a Bayesian neural network with one layer of 50 nodes and Rectified Linear Units (ReLU) activations, and fix the observational noise to $\sigma_y = 0.1$. Across all models, we assume the same priors $p(\W) = p(\z) = p(\decp) \doteq \mathcal{N}(0,0.1)$.
We perform cross-validation of the step size $\lambda_1 \in \{0.1, 0.01, 0.001, 0.0001 \}$ and fix the batch-size to 128. % \in \{16, 128, 512 \}$.
We use Adam~\citep{kingma2014adam} as the optimizer and the joint unnormalized posterior distribution $p( \W | \mathcal{D}_{train})$ as the objective function.

For the proposed approach proj-BNN, we gather 150 weight samples with FGE to train the prediction-constrained autoencoder. The architecture of the projection function is cross-validated between either 1 layer of 20 nodes, or 2 layers of 10 nodes. Open source code is available in the public repository: \emph{Anonymized}. Further details on the experimental setup and baselines can be found in the Appendix. 

%%%%   FIGURE
%\begin{figure*}[h!]
%	\includegraphics[width=0.5\textwidth]{./figs/overall/boston.png}\includegraphics[width=0.5\textwidth]{./figs/overall/concrete.png}
%	\includegraphics[width=0.5\textwidth]{./figs/overall/energy.png}\includegraphics[width=0.5\textwidth]{./figs/overall/power_plant.png}
%	\includegraphics[width=0.5\textwidth]{./figs/overall/wine_red.png}\includegraphics[width=0.5\textwidth]{./figs/overall/wine_white.png}
%	\includegraphics[width=0.5\textwidth]{./figs/overall/yacht.png}\includegraphics[width=0.5\textwidth]{./figs/overall/naval.png}
%	\caption{Test log-likelihood for UCI benchmark datasets for best dimensionality of $z$-space (see Figure~\ref{fig:var_dim_u} for performance across different dimensionality of the latent space $D_{z}$). Red dotted horizontal line corresponds to Proj-BNN performance (our approach).
%		Baselines methods are: 1) BBB: mean field (Blundell, et.al 2015); 2) MNF: multiplicative normalizing flow (Louizos et.al, 2017); 3) MVG: multivariate Gaussian prior BNN (Louizos et.al, 2016). 
%		Ablations of Proj-BNN are: Proj-BNN with linear projections (linear), Proj-BNN without training the autoencoder, i.e., only stage 3 in Alg.~\ref{algo:framework} (1-stage), Proj-BNN modeling uncertainty only in $\z$ ($q(\z)$-only).
%		\textbf{In all but two cases Proj-BNN performs better or as well as the baselines.}}
%	\label{fig:UCI_testll}
%\end{figure*}
%%%%
%%%   FIGURE
%%%

%\paragraph{Inference in latent space can improve model generalization.}
\textbf{Generalization in benchmark datasets.}
 Figure~\ref{fig:UCI_testll} shows that inference in latent space can improve model generalization. On datasets where ground truth distributions are not available for comparison and the inferred distributions are not easily visualized, we argue that the higher quality posterior and posterior predictive potentially obtained from Proj-BNN can be observed through an improvement in the generalizability in the models we obtain. We compare the generalization performance, measured in terms of marginal test likelihood, of Proj-BNN with the three baseline models, BBB, MNF and MVG. In Figure \ref{fig:UCI_testll}, we see that Proj-BNN performs competitively, if not better than baseline models, on all but one dataset.

%\paragraph{Clinical application.}
%XXX
%%%%   FIGURE
%\begin{figure}[h!]
%	\includegraphics[width=0.5\textwidth]{./figs/overall/boston.png}
%	\caption{Test loglikelihood for Intensive Care Unit real-world dataset. Baselines: a) BBB: mean field (Blundell, et.al 2015); b) MNF: multiplicative normalizing flow (Louizos et.al, 2017); c) MVG: multivariate Gaussian prior BNN (Louizos et.al, 2016)}
%\end{figure}
%%%%

%% file: 05_conclusion.tex
\section{Discussion}\label{sec:discussion}
\textbf{The geometry of weight posteriors impacts the quality of variational approximations.}
Experimental results on the synthetic data demonstrates the advantage of Proj-BNN in cases where the geometry of the true posterior over weights is complex, e.g. multimodal or highly non-convex. Here, traditional mean field variational inference will tend to capture only a small part of the true posterior, e.g. a single mode or a small convex region, due to the zero-forcing nature of KL-divergence minimization. Furthermore, in these cases, we find that optimization for variational inference tend to be easily trapped in undesirable local optima. In contrast, provided with a robust non-linear projection onto a low dimensional latent space, we are able to drastically reduce the complexity of the optimization problem, e.g. by reducing the number of parameters to be optimized. As a result, the posterior predictive distributions obtained by Proj-BNN often capture more uncertainty, whereas comparable methods that perform inference directly on weights tend to underestimate uncertainty and, thus, can produce over-confident predictions.

On real datasets, Proj-BNN outperforms or remains competitive, in terms of model generalization, with comparable baseline methods while working with representations of significantly smaller dimensionality. This is again evidence that performing inference in lower dimensional latent space can better capture complex posterior distributions in weight space. In cases where Proj-BNN underperforms in comparison to the baselines, we conjecture that the shortcoming may be due to insufficient sampling of the weight space during the first stage of training with fast geometric ensembling.

\textbf{The quality of the non-linear latent projection impacts the quality of variational inference.} 
The quality of the variational approximations obtained by Proj-BNN relies on 1) the ability to characterize the set of plausible neural network weights given a data set and 2) the ability to learn informative transformations between latent space and weight space.
The former requires us to sample intelligently from the weight space. We currently use Fast Geometric Ensembling which is sample-efficient, but we see opportunities for future research to incorporate training objectives that explicitly encourage diversity of the samples we obtain.
For condition 2), we note that learning transformations that are able to reconstruct weights from their latent representations is not necessarily helpful for inference, as the reconstructed weights might encode to models that suffer from a drastic decrease in predictive accuracy. We address this by adding a predictive-constrained autoencoder for learning the transformations. Additional constraints may be incorporated here to further improve inference in the latent space.

\section{Conclusion}\label{sec:conclusion}
 
 In this paper, we have presented a framework, Proj-BNN, for performing approximate inference for Bayesian Neural Networks that avoid many of the optimization problems of traditional inference methods. In particular, we are able to better capture the geometry of the posterior over weights by learning a probability distribution over non-linear transformations of the weight space onto a simpler latent space and perform mean-field variational inference in the latent space.
 
 Using synthetic data sets, we show that, in cases where the posterior over weights exhibits complex geometry (e.g. is multimodal), variational inference performed on \textit{weights} space will often become trapped in local optima, whereas variational distributions in \textit{latent} space can more easily capture the shape of the true posterior through a non-linear transformation. We compare Proj-BNN with five relevant baselines, each of which is an enrichment of the standard variational approach for BNN inference, and show that Proj-BNN is able to better capture posterior predictive uncertainty (without compromising predictive accuracy) on synthetic data. On 8 real datasets, we show that, in terms of test likelihood, Proj-BNN is able to perform competitively if not better than several baselines, working in latent dimensions that are much smaller than the dimensions of the weight space.
 
%We note that further gains in the quality and efficiency of inference can be attained by improving each phase of our framework. That is, one can implement a number of sample-efficient methods for learning the solution set of the regression model in weight space \citep{zhang2015diversity}; one can regularize the latent representation learned by the autoencoder to have desirable geometries~\citep{hosseini2016deep}; one can place complex priors on the latent space, such as nonparametric variational inference~\citep{gershman2012nonparametric}. 

%Future work:
%\begin{itemize}
%	\item Further gains can be obtained by putting more complex priors on the reduced space, e.g., using a full multivariate Gaussian or a non-parametric mixture prior instead of a diagonal multivariate Gaussian prior.
%\end{itemize}